%% file: root.tex
\documentclass[letterpaper, 10 pt, conference]{ieeeconf}  
\IEEEoverridecommandlockouts                              

\usepackage{amsmath,amsfonts,amssymb,bm}
\usepackage{cite}
\usepackage{hyperref}
\usepackage{graphicx}
\usepackage{url}
\usepackage{colortbl}
\usepackage{color,xcolor}
\usepackage{caption}
\usepackage{float} 
\usepackage{subfig}
\usepackage{subcaption} 
\usepackage{booktabs}
\usepackage{adjustbox} 
\usepackage{pifont} 

\title{\LARGE \bf 
Is Intermediate Fusion All You Need for UAV-based Collaborative Perception?
}
\author{Jiuwu Hao$^{1}$, Liguo Sun$^{2}$, Yuting Wan$^{1}$, Yueyang Wu$^{1}$, Ti Xiang$^{1}$, Haolin Song$^{1}$ and Pin Lv$^{2*}$
\thanks{*Corresponding author}
\thanks{$^{1}$Jiuwu Hao, Yuting Wan, Yueyang Wu, Ti Xiang and Haolin Song are with School of Artificial Intelligence, University of Chinese Academy of Sciences, and with the Key Laboratory of Cognition and Decision Intelligence for Complex Systems, Institute of Automation, Chinese Academy of Sciences}
\thanks{$^{2}$Liguo Sun and Pin Lv are with the Key Laboratory of Cognition and Decision Intelligence for Complex Systems, Institute of Automation, Chinese Academy of Sciences}
}

\begin{document}
\maketitle
\thispagestyle{empty}
\pagestyle{empty}

\input{sec/0_abstract}

\input{sec/1_intro}

\input{sec/2_relatedwork}
\input{sec/3_method}

\input{sec/4_experiments}

\input{sec/5_limitations}

\input{sec/6_conclusion}

\section*{ACKNOWLEDGMENT}
This work was supported by the National Science and Technology Major Project under Grant 2022ZD0116409.

\bibliographystyle{IEEEtran}
\begin{samepage}
\bibliography{reference}
\end{samepage}

\end{document}

%% file: sec/0_abstract.tex
\begin{abstract}
Collaborative perception enhances environmental awareness through inter-agent communication and is regarded as a promising solution to intelligent transportation systems.
However, existing collaborative methods for Unmanned Aerial Vehicles (UAVs) overlook the unique characteristics of the UAV perspective, resulting in substantial communication overhead.
To address this issue, we propose a novel communication-efficient collaborative perception framework based on late-intermediate fusion, dubbed \texttt{LIF}.
The core concept is to exchange informative and compact detection results and shift the fusion stage to the feature representation level.
In particular, we leverage vision-guided positional embedding (VPE) and box-based virtual augmented feature (BoBEV) to effectively integrate complementary information from various agents.
Additionally, we innovatively introduce an uncertainty-driven communication mechanism that uses uncertainty evaluation to select high-quality and reliable shared areas.
Experimental results demonstrate that our \texttt{LIF} achieves superior performance with minimal communication bandwidth, proving its effectiveness and practicality.
Code and models are available at \href{https://github.com/uestchjw/LIF}{\textcolor{purple}{https://github.com/uestchjw/LIF}}.
\end{abstract}

%% file: sec/1_intro.tex
\section{INTRODUCTION}
Vision-centric 3D object detection is a fundamental requirement for modern transportation systems, which helps robotic agents to achieve accurate and robust understanding of the surrounding environment. 
Perception system based on Unmanned Aerial Vehicles (UAVs), offering broader sensing range and enhanced deployment flexibility compared to vehicle- or roadside-based solutions, has attracted rising attention from both academia and industry \cite{UAV_Meet_LLMs, MM-Tracker, PRL-Track}.
To address the limitations of single-agent perception, collaborative perception has emerged as a new paradigm that integrates complementary information from multiple agents to improve environmental awareness \cite{DAIR-V2X, MRCNet}.
Therefore, multi-UAV collaborative perception holds significant potential for advancing intelligent transportation systems.
\par
Communication efficiency is essential in practical collaborative systems.
Most existing multi-UAV collaborative perception methods focus on intermediate fusion strategies \cite{DHD, UAV3D, UVCPNet, UCDNet, Griffin}, which originate from the autonomous driving domain, and involve transmitting and fusing neural features to optimize the performance-bandwidth trade-off.
However, these methods overlook the unique characteristics of UAV platforms, resulting in unnecessary communication overhead.
First, compared to vehicle-mounted or roadside devices, the limited communication capacity and computational resources of UAVs impose stricter constraints on transmission bandwidth.
While extensive works have been devoted to improving communication efficiency \cite{CodeFilling, Where2comm, Who2com, When2com, Which2comm}, the high-dimensional nature of intermediate features inevitably leads to substantial bandwidth consumption.
Second, operating at higher altitude with reduced occlusion, UAVs provide more accurate predictions than ground-based perspectives. 
This makes efficient collaboration based on the exchange of detection results a feasible solution, as the outputs already contain sufficient and reliable scene information.
Thus, the question is raised: \textit{Is intermediate fusion all we need for UAV-based collaborative perception?}
\par
\begin{figure*}[thpb]
  \centering
  \includegraphics[width=\linewidth, keepaspectratio]{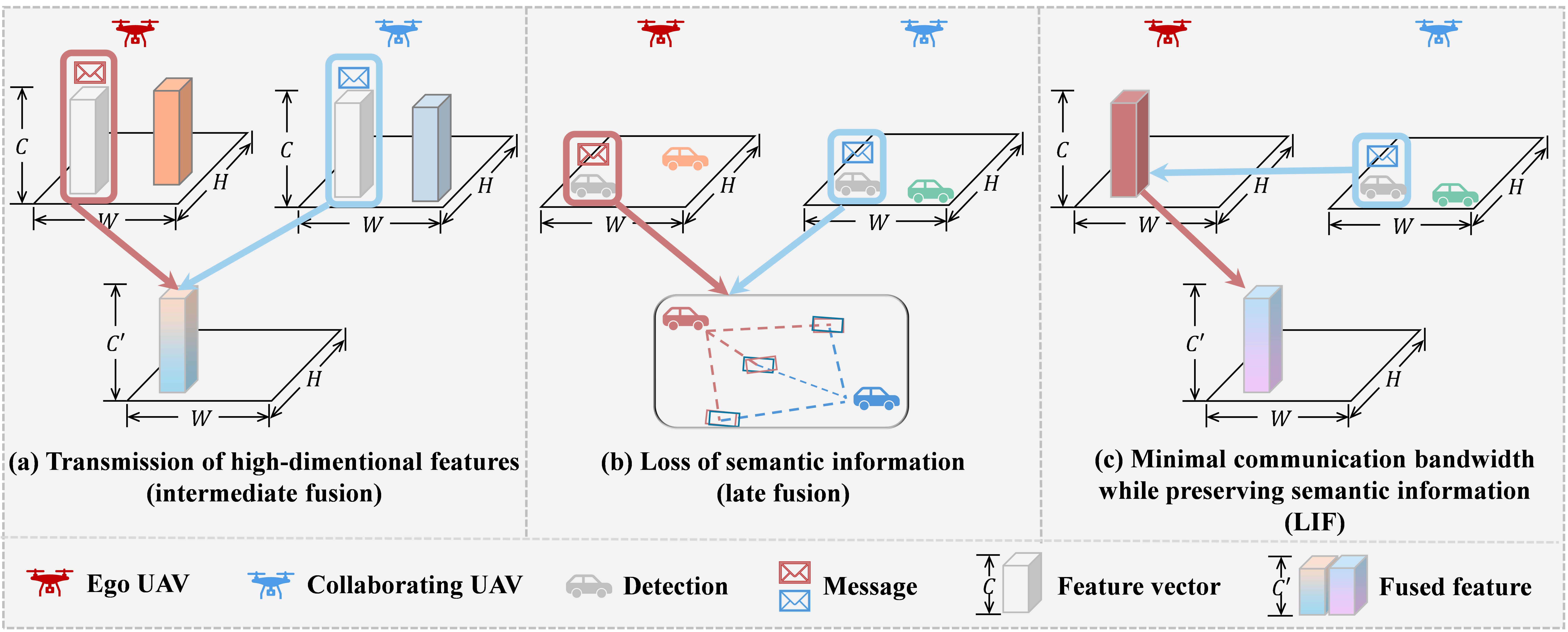}
  \caption{\textbf{Comparison of various collaborative strategies.} (a) Intermediate fusion exchanges and aggregates high-dimensional neural features among agents, resulting in substantial communication overhead. (b) Late fusion focuses on integrating detection results directly, which leads to the scene information loss and degraded performance. (c) Our proposed late-intermediate fusion (\texttt{LIF}) shares compact prediction outputs and incorporates them into the ego-agent's BEV features, effectively reducing communication redundancy while maintaining perception performance.}
  \label{figure: shiyitu}
\end{figure*}

In this work, we propose a novel late-intermediate fusion framework, dubbed \texttt{LIF}, to reduce communication redundancy in collaborative perception.
The core idea is to transmit compact perception results among collaborating agents and integrate them with the ego-agent's intermediate features through a learnable approach; see Fig. \ref{figure: shiyitu}.
By shifting the fusion stage, we mitigate the information loss and performance degradation commonly associated with late fusion.
Moreover, each agent transmits only predictions, alleviating the heavy communication overhead of intermediate fusion.
\par
Specially, \texttt{LIF} consists of three key components: i) Vision-guided positional embedding VPE, which leverages 2D perception results to refine spatial attention, enabling the model to focus more on critical regions; ii) Box-based virtual augmented feature BoBEV, which effectively integrates complementary information from other agents into the Bird's-Eye-View (BEV) representation; iii) Uncertainty-driven communication, which filters high-quality and trustworthy perception outputs through uncertainty estimation.
To validate the effectiveness of our \texttt{LIF}, we conduct extensive experiments on the multi-UAV collaborative perception dataset UAV3D \cite{UAV3D}.
Experimental results show that \texttt{LIF} outperforms state-of-the-art (SOTA) methods with minimal communication overhead, achieving 72.1\% mAP and 61.4\% NDS.
\par
In summary, our contributions are as follows:
\begin{itemize}
    \item We present a novel late-intermediate fusion framework for multi-UAV collaborative perception, namely \texttt{LIF}, which significantly reduces communication redundancy while maintaining detection performance.
    \item We propose two feature enhancement modules, vision-guided positional embedding VPE and box-based virtual augmented feature BoBEV, which facilitate the effective integration of complementary information.
    \item We design a novel uncertainty-driven communication mechanism, which prioritizes high-quality and reliable perception outputs during information exchange process.
    \item Extensive experiments show that \texttt{LIF} achieves superior performance with minimal communication bandwidth, demonstrating its feasibility in real-world collaborative scenarios.
\end{itemize}

%% file: sec/2_relatedwork.tex
\section{RELATED WORKS}
\textbf{Collaborative Perception.}
As an emerging perception paradigm of multi-agent cooperation, collaborative perception has attracted widespread attention \cite{OPV2V, MRCNet, V2X-PC, CMiMC}.
Collaborative perception can be categorized into early fusion, intermediate fusion, and late fusion based on the cooperation stage.
Early fusion \cite{Cooper} transmits raw sensor data, such as images or point clouds, which preserves the most complete perception information but consumes a large amount of transmission bandwidth.
Late fusion \cite{DAIR-V2X} shares each agent's perception results, using minimal communication volume, but can lead to a degradation in perception performance.
Most prevailing works explore intermediate fusion \cite{CoBEVT, CoCMT, HEAL}, where implicit features from intermediate layers are exchanged among collaborating agents, aiming to achieve a trade-off between performance and bandwidth.
Recently, several studies \cite{UAV3D,DHD,UCDNet,UVCPNet} introduce UAVs into collaborative perception system in urban traffic scenarios.
Similarly, in this work, we focus on multi-UAV collaboration and propose a pragmatic and robust cooperative framework to enhance communication efficiency.

\par
\textbf{Communication-efficient Collaboration.}
Due to the stringent limitation on transmission capacity in real-world applications, previous studies are dedicated to developing efficient communication strategies to reduce bandwidth usage.
Traditional intermediate fusion methods that transmit BEV feature maps primarily reduce transmission overhead through the selection of collaborators \cite{Who2com, When2com}, spatial filtering \cite{Where2comm, CoCa3D, How2comm, What2comm}, and compression of message representations \cite{V2X-ViT, CoBEVT, CodeFilling}.
Another branch of methods reduces communication redundancy by transmitting instance-level query features \cite{QUEST, CoCMT}.
However, operating at the feature level inherently leads to the transmission of high-dimensional information, which inevitably incurs significant communication cost.
A contemporaneous work, Which2comm \cite{Which2comm}, proposes leveraging bounding-box-based object-level sparse features, but it still transmits redundant hidden features.
The work most closest to our approach is CoSDH \cite{CoSDH}, which introduces a late-intermediate hybridization mode to address the challenge of limited communication bandwidth. 
However, it does not fully integrate the detection results between agents.
In contrast to these prior works, we propose transmitting only detection results and introduce a novel fusion method to effectively extract and integrate semantic information among collaborators.

%% file: sec/3_method.tex
\section{METHODOLOGY}
\begin{figure*}[t]
  \centering
  \includegraphics[width=\textwidth]{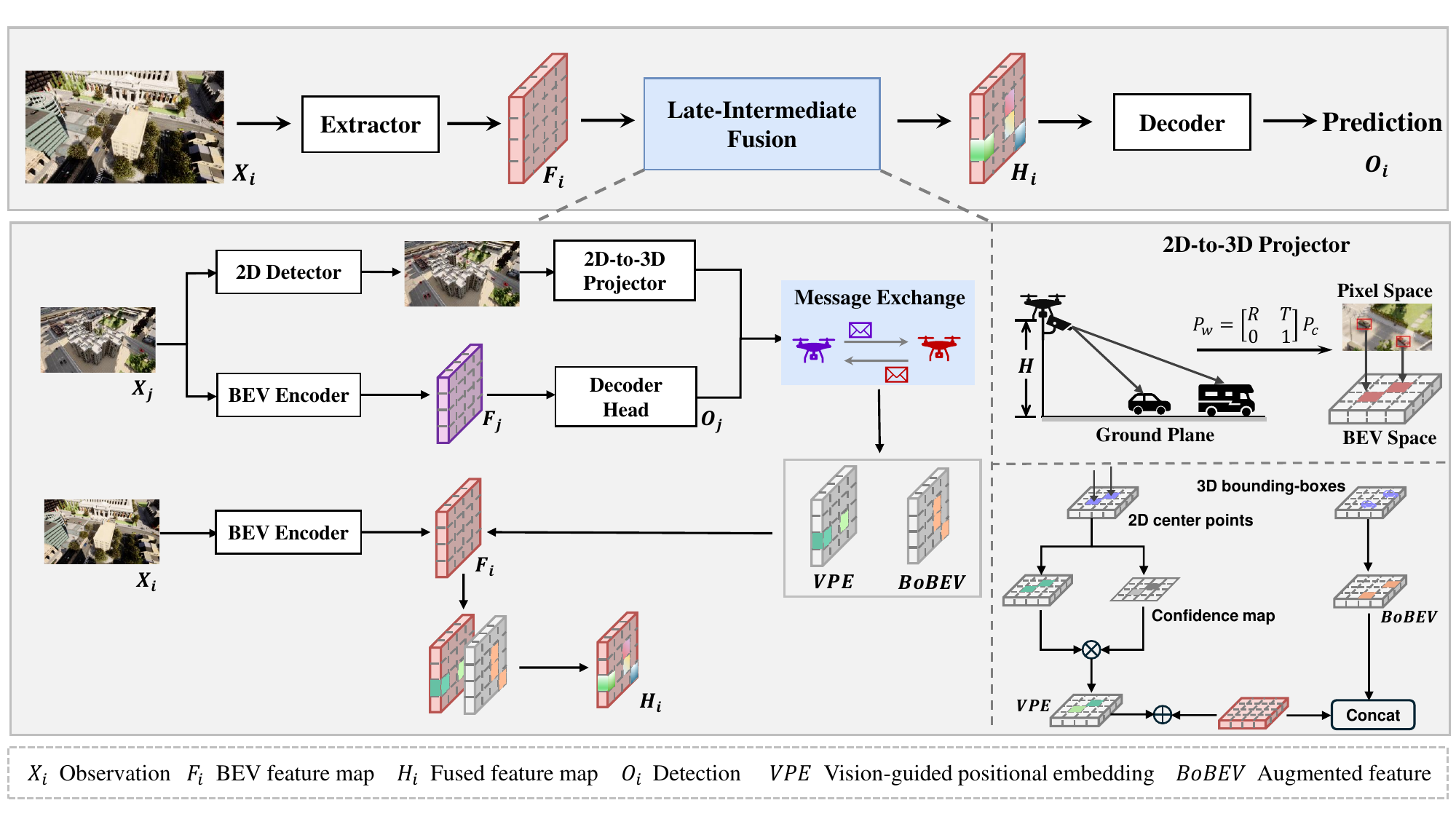}
  \caption{\textbf{The overall architecture of the proposed \texttt{LIF}.} Each agent performs independent environment perception and generates both 2D and 3D detection results. A geometry-prior-based 2D-to-3D projector is utilized to transform the 2D predictions into the 3D space. Then, the uncertainty-driven communication facilitates information exchange among collaborating agents. We refine the ego-agent's feature map by incorporating the vision-guided positional embedding VPE and the box-based virtual augmented feature BoBEV. Finally, the decoder head accepts the fused features as input and produces the 3D collaborative detection outputs.}
  \label{figure: overview}
\end{figure*}

In this work, we focus on improving the communication efficiency of multi-UAV collaborative perception in complex scenarios.
To achieve this, we propose a pragmatic late-intermediate fusion framework, \texttt{LIF}. 
To facilitate the explanation, we introduce \texttt{LIF} in the context of camera-based 3D object detection and assume no localization errors in the collaboration system.

\subsection{Overview}
Fig. \ref{figure: overview} shows the overall architecture of our proposed \texttt{LIF}. First, each agent independently performs environmental perception and generates both 2D and 3D detection results. A geometry-based 2D-to-3D projector is used to transform the 2D object points from pixel space to BEV space. Then an uncertainty-driven communication is employed between collaborating agents to exchange perception results and pose information. Upon receiving the informative messages from other agents, the ego-agent integrates these cues into its own features through a late-intermediate fusion module. The decoder head takes the fused features as input and outputs the final 3D detection outputs.

\subsection{Individual Perception}
Given $N$ UAVs in the scene, each equipped with $M$ surrounding cameras, every agent independently perceives the environment based on its own raw input, producing both 2D and 3D predictions. For the 2D perception, given the RGB images $X_i^{(t)}$ of $i$-th agent at timestamp $t$, the 2D detection results are obtained as $D_i^{(t)} = {\Theta _i}(X_i^{(t)}) \in {\mathbb{R}^{K \times 5}}$, where ${\Theta _i}( \cdot )$ is the 2D detector used by $i$-th agent and $K$ is the number of objects in the pixel space. Each 2D target $b_{2d}=(x,y,w,h,s)$ consists of the center coordinate $(x,y)$, box size $(w,h)$, and confidence score $s$. Here, we choose RF-DETR \cite{rf-detr} as the detector for its real-time performance.
\par
For the 3D perception pipeline, BEVFusion \cite{BEVFusion} is selected as the backbone to extract BEV features as $F_i^{(t)} = {\Phi _{enc}}(X_i^{(t)}) \in {\mathbb{R}^{H \times W \times C}}$, where $H,W,C$ are the height, width and channel of the feature map. The detection head outputs the 3D perception results of individual agent by $\widehat D_i^{(t)} = {\widehat \Theta _i}(F_i^{(t)}) \in {\mathbb{R}^{\widehat K \times 8}}$, where $\widehat \Theta ( \cdot )$ is the 3D output header and $\widehat K$ is the number of objects in the BEV space. The 3D bounding box $b_{3d}$ is an 8-dimensional vector $[x,y,z,w,h,l,\theta,s]$, where $(x,y,z)$ represent the center location, $(w,h,l)$ represent the size, $\theta$ represents the heading angle, and $s$ is the confidence score. Note that we focus solely on the detection outcomes, permitting connected agents to employ distinct models, enabling effective collaboration in heterogeneous scenarios.
\par
Vanilla BEVFusion \cite{BEVFusion} is tailored for the vehicle-mounted perspective in autonomous driving, where the smallest feature map is downsampled to $\frac{1}{8}$ of the original image size. Due to the typically smaller size of targets in UAV images, we double the feature map size while reducing the channel dimension to avoid increasing computational overhead.

\subsection{Coordinate Transformation}
\label{subsection: coordinate transformation}
After obtaining the detection results in the 2D image, we utilize the camera pose information provided by the UAV to determine the object's position in the world coordinate system. 
\par
In the camera coordinate system, the X-axis points to the right of the camera's optical center, the Y-axis points downward, and the Z-axis points forward along the camera's viewing direction. The transformation from the 2D pixel coordinate $(u,v)$ to the 3D camera coordinate $\left( {{x_c},{y_c},{z_c}} \right)$ is determined by the camera intrinsic matrix, which is formulated as:
\begin{equation}
    \left[ \begin{gathered}
      u \hfill \\
      v \hfill \\
      1 \hfill \\ 
    \end{gathered}  \right] = \frac{1}{{{z_c}}}\left[ 
    \begin{matrix}
      f_x & 0 & c_x \\ 
      0 & f_y & c_y \\ 
      0 &  0  & 1   \\
    \end{matrix}  \right] \cdot \left[ 
    \begin{gathered}
      {x_c} \hfill \\
      {y_c} \hfill \\
      {z_c} \hfill \\ 
    \end{gathered}  \right]     
\end{equation}
where ${f_x},{f_y}$ are the focal lengths of the camera, and ${c_x},{c_y}$ are the center coordinates in the image plane. 
\par
The equation of the ray corresponding to the pixel point $(u,v)$ in the camera coordinate can be expressed as:
\begin{equation}
    P\left( r \right) = r \cdot \left( {\frac{{u - {c_x}}}{{{f_x}}},\frac{{v - {c_y}}}{{{f_y}}},1} \right)
\label{equ-2}
\end{equation}
where $r$ is the scalar parameter.
\par
Based on the camera's pose and altitude, the transformation from world coordinate ${P_w}$ to camera coordinate ${P_c}$ is as follows:
\begin{equation}
    {P_w} = \left[ {\begin{matrix}
                  R&T \\ 
                  0&1 \\
                \end{matrix}} \right] \cdot {P_c}  
\end{equation}
where $R$ is a 3x3 rotation matrix, $T$ is a 3x1 translation vector. Then we transform the line equation \eqref{equ-2} into the world coordinate system, which is formulated as:
\begin{equation}
    {P_w}\left( r \right) = T + r \cdot R \cdot \left( {\frac{{u - {c_x}}}{{{f_x}}},\frac{{v - {c_y}}}{{{f_y}}},1} \right)
    \label{equ-4} 
\end{equation}
\par
In 2D images, the objects of our interest, such as vehicles and pedestrians, typically maintain a relatively constant height above the ground. Given the object height $z_w=h$, the coordinates $x_w$ and $y_w$ of the object in the world coordinate system can be obtained using \eqref{equ-4}.
\par

\subsection{Uncertainty-driven Communication}
We adopt an uncertainty-driven communication mechanism to select high-quality, reliable detection results, which serve as complementary information to provide a more comprehensive scene representation for the ego agent.
Specifically, for the $i$-th agent, its uncertainty map ${U_i} \in {\mathbb{R}^{H \times W \times 1}}$ is calculated as:
\begin{equation}
{U_i} = 1 - \left| {{S_i} - {\phi _i}} \right|
\end{equation}
where ${S_i} \in {[0,1]^{H \times W}}$ refers to the classification output of $i$-th agent and ${\phi _i}$ is the detection threshold. The demand map that determines which position $j$-th agent will share with $i$-th agent is:
\begin{equation}
{R_{ij}} = {U_i} \odot (1 - {U_j})
\end{equation}
where $\odot$ is the element-wise multiplication.
The $j$-th agent will transmit the decoder output of positions with ${R_{ij}}$ exceeds the demand threshold ${\phi _{dem}}$.
\par
\textbf{Comparison against Where2comm.} It is important to distinguish our method from the previous communication approach, Where2comm \cite{Where2comm}, to clarify the differences.
Both works employ a spatial selection strategy to prioritize the transmission of high-value locations.
Differently, Where2comm relies solely on objectness as the evaluation criterion, neglecting the uncertainty inherent in the model's predictions.
Instead, our method considers both objectness and uncertainty, choosing high-value and trustworthy detection results, where foreground areas generally exhibit a higher demand value due to the use of a small detection threshold.
\par
Moreover, we argue that high-confidence background outputs also hold transmission value, as they can mitigate false positives in the ego-agent's predictions.
Thus, as communication bandwidth increases further, we prioritize sharing the highly certain background regions (those uncertain for the ego-agent), rather than ambiguous and misleading locations.

\subsection{Late-intermediate Fusion}
We argue that the detection results already encapsulate sufficient scene information for UAV-based collaboration. Therefore, we transmit and share individual predictions among collaborators, rather than the conventional redundant neural features. We integrate the received perception outputs into the ego-agent's BEV features in a learnable method, enabling effective interaction between agents. It is worth noting that, in addition to 3D bounding boxes, we also leverage 2D detection results to compensate for the limitations of the 3D single-agent perception model.
\par
\textbf{Vision-guided positional embedding VPE.}\
We use the 2D detection results from other agents and coordinate transformations to guide the ego-agent in focusing on specific locations in the BEV feature map.
Let ${P_w} = \left\{ {{{\left( {x,y,z} \right)}_k}} \right\}$ denote the world coordinates of the target center points from the image detection results. Since the ego coordinate system and the world coordinate system share the same axis directions, we can obtain the target coordinates ${P_{ego}}$ in the ego coordinate system by applying a translation vector $T_{ego2w}$, which is formulated as:
\begin{equation}
    {P_{ego}} = {P_w} - {T_{ego2w}}
\end{equation}
\par
And given the resolution $\delta$ of the BEV space, we obtain the position of the target point in the BEV coordinate system as ${P_{bev}} = {P_{ego}}/\delta$ (ignoring the vertical direction). 
\par
We adopt a learnable approach to generate the positional embedding, where non-zero positions indicate the presence of target points $P_{bev}$. The vision-guided positional embedding $VPE \in {\mathbb{R}^{H \times W \times C}}$ at location $(m,n)$ is:
\begin{equation}
    VP{E_{m,n}} = \left\{ \begin{gathered}
      Q \in {\mathbb{R}^C},{\text{   }}(m,n){\text{ in }}{P_{bev}} \hfill \\
      0,{\text{           else}} \hfill \\ 
    \end{gathered}  \right.
\end{equation}
where $Q$ is the learned positional feature, and $C$ is the feature dimension.
\par
\begin{figure*}[t]
  \centering
  \includegraphics[width=\linewidth, keepaspectratio]{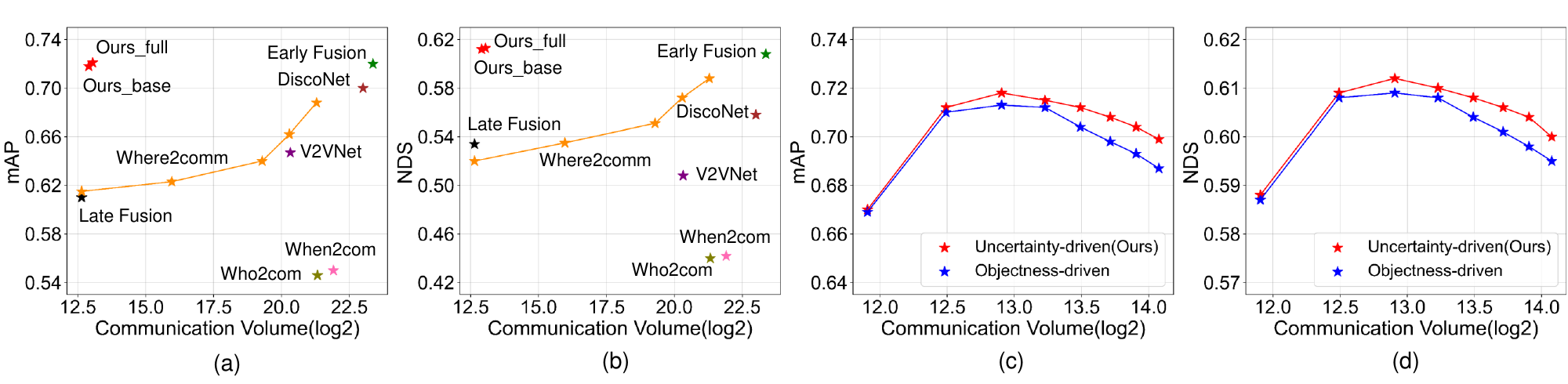}
  \caption{\textbf{Accuracy-bandwidth trade-off.} (a)-(b) illustrate the performance comparison between our \texttt{LIF} and previous collaborative methods. (c)-(d) present the comparison between the proposed uncertainty-driven communication and the conventional objectness-driven communication. ``Ours\_base" refers to using BoBEV alone for feature augmentation, while ``Ours\_full" leverages both VPE and BoBEV modules.}
  \label{figure:communication_volume}
\end{figure*}
Then, we enhance the positional embeddings based on the confidence of the detection results, which is:
\begin{equation}
    VPE' = S \odot VPE
\end{equation}
where $S$ is the confidence map generated by 2D detector.
\par
Finally, the vision-guided positional embedding is added to the BEV feature map, indicating that the corresponding positions should receive more attention, as formulated below:
\begin{equation}
    {F_i^{'}} = {F_i} + VPE',{\text{  }}{F_i^{'}} \in {\mathbb{R}^{H \times W \times C}}
\end{equation}
where $F_i^{'}$ is the refined feature map, $F_i$ is the original BEV feature and $VPE'$ is the generated positional embedding.
\par
\textbf{Virtual augmented feature BoBEV.} Although the UAV's perspective offers less semantic ambiguity, directly fusing detection results (i.e., 3D bounding boxes) between connected-UAVs still yields suboptimal performance. Our goal is to achieve more thorough and comprehensive information integration, minimizing the impact of information loss caused by merely transmitting detection results.
\par
Inspired by MoDAR \cite{MoDAR}, we propose integrating the 3D bounding boxes into the processing pipeline of the ego-UAV at an earlier stage, enabling the model to learn how to utilize complementary information from other UAVs more effectively.
Unlike MoDAR, which directly integrates virtual modalities into LiDAR point clouds, embedding 3D bounding boxes into 2D images is not a trivial task. 
It is also worth noting that integrating perceptual information from other agents at earlier stages tends to introduce greater processing latency. 
Therefore, we choose to integrate the received 3D bounding boxes into the ego-UAV's BEV features, referred to as the Box-based augmented BEV feature map (BoBEV).
\par
Given a set of received 3D bounding boxes $B = \left\{ {{b_{3d}^{(k)}}} \right\}$, each box $b_{3d} = \left( {x,y,z,w,h,l,\theta,s } \right)$, we use the geometric feature and confidence score of the boxes to form the augmented feature map $BoBEV \in {\mathbb{R}^{H \times W \times 5}}$, as expressed by:
\begin{equation}
    BoBE{V_{m,n}} = \left\{ \begin{gathered}
      {b_{3d}^{whl\theta s}},{\text{   }}(m,n){\text{ in }}{P_{bev}} \hfill \\
      0,{\text{        else}} \hfill \\ 
    \end{gathered}  \right.
\end{equation}
where ${b_{3d}^{whl\theta s}}$ is composed of the bounding box's size, heading direction and confidence score.
Then, we concatenate $BoBEV$ with the features after adding positional embedding to obtain the final fused feature map ${H_i} \in {\mathbb{R}^{H \times W \times (C + 5)}}$, which is formulated as:
\begin{equation}
    {H_i} = Concat\left( {\left[ {F_i^{'},BoBEV} \right]} \right)
\end{equation}

\subsection{Detection Head}
    Same as \cite{BEVFusion}, we adopt a center heatmap head to predict the center location of bounding boxes and different regression heads to estimate the size, heading and velocity of objects. We use the Gaussian focal loss \cite{Gaussianfocalloss} for classification and $L_1$ loss for regression.

%% file: sec/4_experiments.tex
\section{EXPERIMENT}
In this section, we first introduce our experimental setup, including the dataset and implementation details. Then, we evaluate the quantitative and qualitative performance of collaborative 3D detection task. Finally, we conduct ablation studies to validate the effectiveness of each proposed module.

\subsection{Experiment Setup}
\textbf{Dataset.} We conduct experiments on the public UAV collaborative perception dataset UAV3D \cite{UAV3D}, which is a large-scale simulated dataset collected by CARLA \cite{CARLA} and AirSim \cite{AirSim}. UAV3D comprises a total of 1,000 scenes, with each scene containing 20 frames. We utilize 14,000 samples for the training set, 3,000 samples for the validation set, and 3,000 samples for the test set.
\par
\begin{figure*}[thpb]
  \centering
  \includegraphics[width=\textwidth]{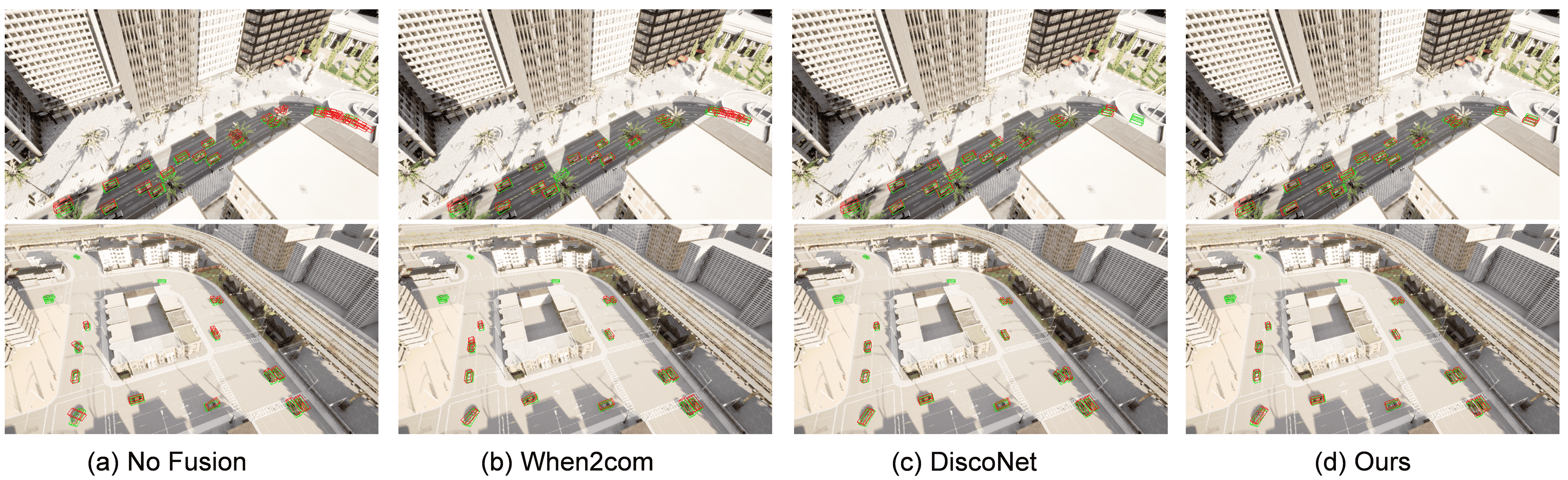}
  \caption{\textbf{Visualization of detection results on the UAV3D dataset.} \textcolor{green}{Green} and \textcolor{red}{red} boxes denote ground-truths and predictions, respectively. It can be observed that our \texttt{LIF} achieves promising detection performance with minimal bandwidth consumption.}
  \label{figure: visualization_overall}
\end{figure*}
\textbf{Implementation details.} Following \cite{UAV3D}, we map all vehicle categories to a single "car" and assume a vehicle height $h$ of $1.5m$. In each frame, 5 UAVs are selected as the collaborating agents and the image size is $(704, 256)$. The detection threshold ${\phi _i}$ is set to 0.15, and the demand threshold ${\phi _{dem}}$ is dynamically adjusted based on the available communication capacity. We set the perception range to $\left[ { - 102.4m,102.4m} \right] \times \left[ { - 102.4m,102.4m} \right]$ for the X-axis and Y-axis in the ego coordinate system. All the models are trained on 4 NVIDIA RTX A6000 GPUs with a batch size of 1 for a total of 24 epochs.
\par
\input{tabs/overall_performance}
\textbf{Communication volume.} Apart from pose information, our proposed \texttt{LIF} only shares detection results among collaborating agents. Specifically, the communication overhead is calculated as ${\log _2}(K \times 3 \times 32/8 + \widehat K \times 8 \times 32/8)$, where $K$ and ${\widehat K}$ denote the number of objects in the 2D and 3D domain, 32 represents the float32 data type and dividing by 8 is used to convert bits into bytes. Here, we transmit the transformed center coordinates and confidence score $(x,y,c)$ for 2D predictions and the full bounding box $(x,y,z,w,h,l,\theta,c)$ for 3D perception results.

\par
\textbf{Evaluation metrics.} We use mAP and NDS as the primary evaluation metrics for 3D object detection task. In addition, we apply three true positive metrics, including mATE, mASE, and mAOE for measuring translation, scale and orientation errors, respectively. The detailed definitions of these metrics can be found in \cite{UAV3D}.

\subsection{Quantitative Evaluation}
\textbf{Main performance analysis.}
We compare the collaborative 3D object detection performance of the proposed \texttt{LIF} with previous methods, as shown in Table \ref{table: overall_performace}. Experimental results demonstrate that \texttt{LIF} achieves the highest accuracy on the UAV3D \cite{UAV3D} dataset, outperforming several mainstream intermediate-fusion methods, such as When2com \cite{When2com}, Who2com \cite{Who2com}, V2VNet \cite{V2Vnet}, and DiscoNet \cite{DiscoNet}. Note that our \texttt{LIF}, which adopts a late-intermediate fusion strategy, even surpasses early fusion that transmits raw data. This is attributed to the enlargement of the feature map size, and a more detailed discussion will be provided in the ``Ablation Studies" part.
\par
\input{tabs/table_Heterogeneity-friendly}
Interestingly, the naive late-fusion approach achieves competitive performance, outperforming some intermediate-fusion methods. Our base model (Ours\_base), which only transmits 3D bounding boxes for collaboration, achieve a mAP of 0.718 and a NDS of 0.612. The model utilizing both 2D and 3D predictions (Ours\_full) achieves a mAP of 0.721 and a NDS of 0.614. These results suggest that the detection outcomes from the UAV perspective contain sufficient scene information, and our novel late-intermediate fusion strategy enables effective and efficient collaboration.
\par
\textbf{Accuracy-bandwidth trade-off comparison.}
Fig. \ref{figure:communication_volume}(a) and (b) illustrate the performance-bandwidth trade-off of our proposed \texttt{LIF} and other methods. We can see that as the communication capacity decreases, the perception performance of Where2comm \cite{Where2comm} degrades rapidly. In contrast, our \texttt{LIF} achieves SOTA perception accuracy with minimal communication consumption, near on par with late fusion method, thereby enabling multi-UAV collaboration in bandwidth-limited real-world applications.
\par
We also compare the proposed uncertainty-driven communication mechanism with previous objectness-driven strategy, as shown in Fig. \ref{figure:communication_volume}(c) and (d). We see that, with limited communication capacity, both approaches exhibit similar perception performance, prioritizing the transmission of reliable foreground objects. However, as the bandwidth increases, the accuracy of our uncertainty-driven communication method gradually surpasses that of the conventional objectness-driven approach. This is because our uncertainty-driven communication eliminates low-quality and unreliable detection results through uncertainty assessment, rather than merely transmitting based on the probability of object presence.
\par
\textbf{Heterogeneous compatibility analysis.} Table \ref{table: Heterogeneity-friendly} illustrates a comparison of the heterogeneous compatibility of various collaborative methods, which is defined as the ability to support agents utilizing diverse perception models.
We see that intermediate fusion approaches, due to their reliance on exchanging feature-level representation, are inherently incompatible with heterogeneous collaboration.
Instead, our \texttt{LIF} shares model-agnostic perception results among agents, enabling effective collaboration under heterogeneous configurations and exhibiting broad application potential in practical scenarios.

\begin{figure}[t]
  \centering
  \includegraphics[width=\linewidth]{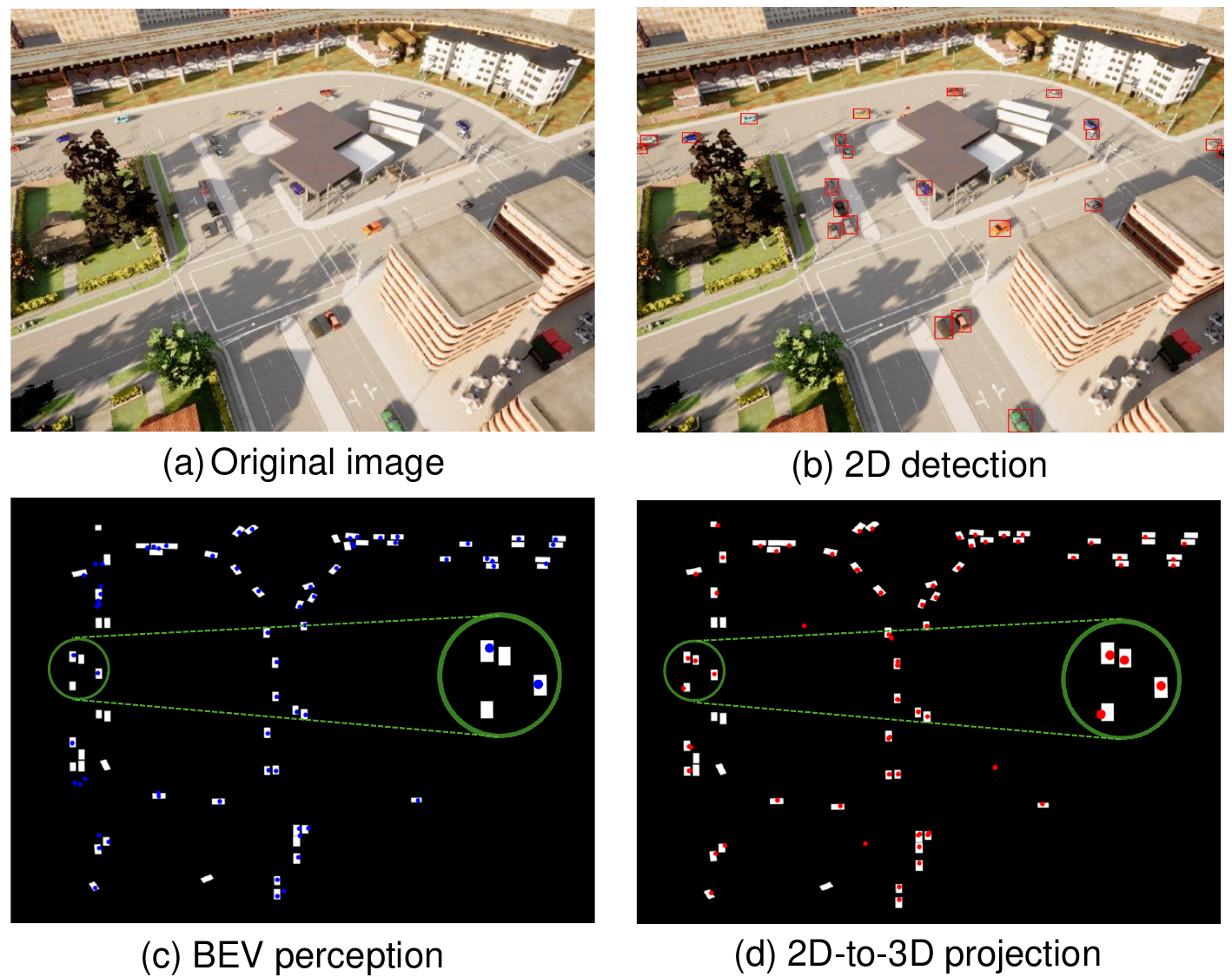}
  \caption{\textbf{Visualization of the 2D-to-3D projection results.} The white boxes represent the ground-truths, while the \textcolor{blue}{blue} and \textcolor{red}{red} points denote the centers of the single-agent BEV perception results and the 2D-to-3D mapping outputs, respectively.}
  \label{figure: 2D-3D Projection}
\end{figure}

\subsection{Qualitative Results}
Fig. \ref{figure: visualization_overall} visualizes the detection results of our proposed \texttt{LIF} compared to other methods on the UAV3D dataset. We see that, despite only transmitting perception outputs among collaborating agents, our method generates more accurate bounding boxes than other approaches. Fig. \ref{figure: 2D-3D Projection} shows the visualization of 2D detection and 2D-to-3D projection. It can be seen that individual BEV perception has its limitations, and the vision-guided positional embedding VPE can help guide the model to focus on the missing foreground regions.
\input{tabs/ablation_study}

\subsection{Ablation Studies}
Table \ref{table: ablation} presents the results of the ablation study on the various proposed modules. Increasing the feature map size (while reducing the channel dimension to maintain computational cost) leads to a 7.9\% improvement in mAP and a 1.0\% improvement in NDS, demonstrating that spatial resolution is more critical for effectively capturing the features of small targets from the UAV perspective.
\par
The incorporation of vision-guided positional embedding VPE can improve the detection performance. However, the absence of detailed bounding box attributes constrains the accuracy of the predicted boxes. The integration of BoBEV, the box-based virtual augmented feature, significantly improves perception performance, enabling efficient fusion of perception results from collaborating agents.

%% file: tabs/overall_performance.tex
\begin{table}[htbp]
\caption{Overall performance on the UAV3D dataset. ``Ours\_base" refers to the use of BoBEV alone, while ``Ours\_full" represents the combination of both VPE and BoBEV to enrich feature representations.}
\label{table: overall_performace}
\begin{center}
\begin{tabular}{c|c c c c c}
\rowcolor{gray!20}
\textbf{Method} & \textbf{mAP$\uparrow$} & \textbf{NDS$\uparrow$} & \textbf{mATE$\downarrow$} & \textbf{mASE$\downarrow$} & \textbf{mAOE$\downarrow$} \\
\hline
\addlinespace[2.5pt]
No Fusion & 0.544 & 0.481 & 0.509 & 0.155 & 0.319 \\ 
Late Fusion & 0.610 & 0.535 & 0.571 & 0.159 & 0.073 \\ 
Early Fusion & 0.720 & 0.608 & 0.391 & \textbf{0.106} & 0.117 \\ 
\midrule
When2com \cite{When2com} & 0.550 & 0.442 & 0.534 & 0.156 & 0.679 \\
Who2com \cite{Who2com} & 0.546 & 0.440 & 0.536 & 0.153 & 0.677 \\
V2VNet \cite{V2Vnet} & 0.647 & 0.508 & 0.508 & 0.167 & 0.533 \\ 
DiscoNet \cite{DiscoNet} & 0.700 & 0.558 & 0.423 & 0.143 & 0.422 \\
\midrule
Ours\_base & 0.718 & 0.612 & \textbf{0.353} & 0.135 & \textbf{0.086} \\ 
Ours\_full & \textbf{0.721} & \textbf{0.614} & 0.355 & 0.137 & 0.090 \\ 
\midrule
\end{tabular}
\end{center}
\end{table}

%% file: tabs/table_Heterogeneity-friendly.tex
\begin{table}[t]
\caption{Comparison of heterogeneous compatibility.}
\label{table: Heterogeneity-friendly}
\begin{center}
\begin{adjustbox}{width=\columnwidth} 
\begin{tabular}{c|c|c|c}
\rowcolor{gray!20}
\textbf{Method} & \textbf{Received message} & \textbf{Fusing level} & \textbf{Heterogeneity-friendly} \\
\hline
\addlinespace[2.5pt]
No Fusion & \ding{55} & \ding{55} & \checkmark \\
Early Fusion & Images/Point clouds & Images/Point clouds & \checkmark \\
Late Fusion & Detection results & Detection results & \checkmark \\
\midrule
When2com \cite{When2com} & Neural features & Neural features & \ding{55} \\
Who2com \cite{Who2com} & Neural features & Neural features & \ding{55} \\
V2VNet \cite{V2Vnet} & Neural features & Neural features & \ding{55} \\
DiscoNet \cite{DiscoNet} & Neural features & Neural features & \ding{55} \\
\midrule
\texttt{LIF} (Ours) & Detection results & Neural features & \checkmark \\

\midrule
\end{tabular}
\end{adjustbox}
\end{center}
\end{table}

%% file: tabs/ablation_study.tex
\begin{table}[t]
\caption{Ablation study of the proposed components on UAV3D dataset. ``LR'' refers to the doubling of the feature map size while reducing the channel dimension.}
\label{table: ablation}
\begin{center}
\begin{adjustbox}{width=\columnwidth} 
\begin{tabular}{c c c| c c c c c}
\rowcolor{gray!20}
\textbf{VPE} & \textbf{BoBEV} & \textbf{LR} & \textbf{mAP$\uparrow$} & \textbf{NDS$\uparrow$} & \textbf{mATE$\downarrow$} & \textbf{mASE$\downarrow$} & \textbf{mAOE$\downarrow$} \\
\midrule
\addlinespace[2.5pt]
 & & & 0.544 & 0.481 & 0.509 & 0.155 & 0.319 \\
 \checkmark & & & 0.637 & 0.478 & 0.396 & 0.160 & 1.547 \\
& \checkmark & & 0.694 & 0.598 & 0.371 & 0.135 & 0.090 \\   
\checkmark & \checkmark & & 0.702 & 0.599 & 0.374 & 0.142 & 0.108 \\
\midrule
& & \checkmark & 0.587 & 0.486 & 0.453 & 0.146 & 0.530 \\
& \checkmark & \checkmark & 0.718 & 0.612 & 0.353 & 0.135 & 0.086 \\ 
\checkmark & \checkmark & \checkmark & 0.721 & 0.614 & 0.355 & 0.137 & 0.090 \\

\midrule
\end{tabular}
\end{adjustbox}
\end{center}
\end{table}

%% file: sec/5_limitations.tex
\section{Limitations}
The perception performance of the proposed \texttt{LIF} is evaluated on the UAV3D dataset \cite{UAV3D}, which is constructed in a simulated environment.
Recently, some real-world UAV collaboration datasets, such as AGC-Drive \cite{AGC-Drive}, have been proposed.
Further experiments are needed to assess the generalizability and robustness of the model in more challenging scenarios.
However, the insight from our experimental analysis has demonstrated the potential of leveraging predicted results for collaboration, where such a simple yet effective method leads to a significant improvement in the performance-bandwidth trade-off.

%% file: sec/6_conclusion.tex
\section{CONCLUSION}
In this paper, we propose \texttt{LIF}, a novel communication-efficient framework for multi-UAV collaboration.
By transmitting compact perception results and integrating complementary information, \texttt{LIF} reduces communication redundancy and enhances detection performance.
Moreover, we present an uncertainty-driven communication mechanism that effectively selects high-quality and reliable perception outputs for sharing.
Experimental results demonstrate that \texttt{LIF} achieves state-of-the-art performance with minimal communication bandwidth.
Further consideration of localization errors will be a future direction.